\documentclass{ecai} 

\usepackage{latexsym}
\usepackage{amssymb}
\usepackage{amsmath}
\usepackage{amsthm}
\usepackage{booktabs}
\usepackage{enumitem}
\usepackage{graphicx}
\usepackage{color}
\usepackage{bm}
\usepackage{algorithm}
\usepackage{algorithmic}
\usepackage{threeparttable}
\usepackage{tabularx}

\newcommand{\BibTeX}{B\kern-.05em{\sc i\kern-.025em b}\kern-.08em\TeX}

\begin{document}


\begin{frontmatter}


\paperid{2123} 


\title{A Meta-Learning Approach for Multi-Objective Reinforcement Learning in Sustainable Home Energy Management}


\author[A]{\fnms{Junlin}~\snm{Lu}\orcid{0000-0002-6014-9419}\thanks{Corresponding Author. Email: J.Lu5@universityofgalway.ie}
}
\author[A]{\fnms{Patrick}~\snm{Mannion}\orcid{0000-0002-7951-878X}
}
\author[A]{\fnms{Karl}~\snm{Mason}\orcid{0000-0002-8966-9100}} 

\address[A]{University of Galway}


\begin{abstract}
Effective residential appliance scheduling is crucial for sustainable living. While multi-objective reinforcement learning (MORL) has proven effective in balancing user preferences in appliance scheduling, traditional MORL struggles with limited data in non-stationary residential settings characterized by renewable generation variations.  Significant context shifts that can invalidate previously learned policies. To address these challenges, we extend state-of-the-art MORL algorithms with the meta-learning paradigm, enabling rapid, few-shot adaptation to shifting contexts. Additionally, we employ an auto-encoder (AE)-based unsupervised method to detect environment context changes. We have also developed a residential energy environment to evaluate our method using real-world data from London residential settings. This study not only assesses the application of MORL in residential appliance scheduling but also underscores the effectiveness of meta-learning in energy management. Our top-performing method significantly surpasses the best baseline,  while the trained model saves 3.28\% on electricity bills, a 2.74\% increase in user comfort, and a 5.9\% improvement in expected utility. Additionally, it reduces the sparsity of solutions by 62.44\%. Remarkably, these gains were accomplished using 96.71\% less training data and 61.1\% fewer training steps.
\end{abstract}

\end{frontmatter}


\section{Introduction}
\label{sec: introduction}
Reducing greenhouse gas emissions has become an essential concern. Electricity production contributes to more than a quarter of the emission \cite{wiki:greenhouse_gas_emissions}, where the residential sector plays a considerable role. It has consumed around 26.8\% of the global electricity in 2022 \cite{wiki_electric_energy_consumption}. Although renewable generations can mitigate this challenge, the intermittent and verifying nature limits their utilization \cite{lu2022multi}. Effective energy management techniques have the potential to overcome this. One of the candidate techniques is reinforcement learning \cite{chauhan2016comparison,real2024optimization}. While the single-objective reinforcement learning (SORL) method is widely used, in residential energy management in practical scenarios the user always needs to make trade-offs between multiple objectives e.g. saving costs and increasing comfort \cite{lu2022multi}. It is therefore more reasonable to render the problem as multi-objective reinforcement learning (MORL) \cite{laber2014set,jalalimanesh2017multi,alegre2023sample}. Given that MORL in residential energy management effectively addresses practical scenarios, it is essential to keep validating the latest MORL method in this field for further improvement. We look into the residential appliance scheduling in this work specifically which is one of the most important parts of energy management.

Alegre et al. recently proposed two MORL algorithms, i.e. \textit{genearalized policy improvement linear support (GPI-LS)} and \textit{genearalized policy improvement prioritized dyna (GPI-PD)} \cite{alegre2023sample}. The agent guarantees rapid training by identifying the corner weight \cite{roijers2016multi} align which the whole policy set can achieve the largest improvement. \textit{GPI-PD} is the first model-based MORL algorithm and \textit{GPI-LS} is the model-free version. However, they were only evaluated in simulated environments in the original work, and it is challenging for them to work well with the non-stationary environment (which arises from intermittent renewable energy production predominantly) of appliance scheduling. In fact, we found in our experiment that the policy trained with one-month data fails to even surpass a simple rule-based policy. 

One intuition is to finetune the policy in response to significant context shifts. However, context shifts are hard to identify due to the absence of explicit labels for these qualitative changes. Furthermore, our experimental findings indicate that merely finetuning a trained policy with new data does not enhance performance effectively. Another intuitive approach is to train the policy with more data, e.g. entire year's data, but this method suffers from the expense of computation overhead. The non-stationary environment nature introduces two critical challenges: \textbf{(i)} With as little data as possible, how can the policy generally be good and easy to finetune? \textbf{(ii)} How to identify the environment context shifts in an unsupervised manner?

Meta-learning  \cite{finn2017model} and \cite{nichol2018first}, is particularly adept at handling scenarios involving the distribution of environment contexts and identifying a parameter initialization that can be rapidly and effectively finetuned to adapt to new contexts. To address the first challenge we have extended the \textit{GPI} algorithms with meta-learning to enable their few-shot finetuning ability. For the second challenge, we have adopted an AE model as an unsupervised method for detecting qualitative shifts in context\cite{audibert2020usad, yang2023adt, yang2024adt}. This approach enables the identification of significant shifts in environmental contexts.

The contributions of this research are as follows:

\textbf{1.} We identify residential energy management as intrinsically compatible with meta-learning and extend \textit{GPI} algorithms with the meta-learning paradigm\footnote{The code and the dataset we used are provided in the supplementary material, we will make the code public in the camera-ready version.}. Our top-performing method significantly surpasses the best baseline,  while the trained model saves 3.28\% on electricity bills, a 2.74\% increase in user comfort, and a 5.9\% improvement in expected utility. Additionally, it reduces the sparsity of solutions by 62.44\%. Remarkably, these gains were accomplished using 96.71\% less training data and 61.1\% fewer training steps. 

\textbf{2.} This study is the first application and evaluation of \textit{GPI-LS/PD} algorithms in residential appliance scheduling and also the first use of a model-based MORL in this field. We have conducted a detailed discussion about model-based MORL in a non-stationary environment.

\textbf{3.} We have developed and released an open-source model for residential energy environments tailored to MORL, which conforms to the standards set by the OpenAI Gym. 

\section{Related Work}
\label{sec: related work}
\subsection{Multi-objective Reinforcement Learning}
\label{subsec: multi-objective reinforcement learning}
MORL methods are designed for multi-objective sequential decision-making problems. The MORL agent is trained to make trade-offs among multiple objectives. The return from the interaction between the agent and the environment in MORL is a vector rather than a scalar as in SORL. Nevertheless, as the return vector is usually scalarized as a utility by calculating the inner product with the preference weight vector \cite{hayes2022practical}, one intuitive way to sort a MORL problem is to separate a MORL policy training from the training of multiple SORL policies based on different preference weight vectors \cite{mannor2001steering, tesauro2007managing,van2013scalarized,roijers2014linear, mossalam2016multi}. This method, however, is computationally expensive when there are many candidate preference weights, or even intractable when the preference is not given. 

This limitation is mitigated by the MORL paradigm conditioned on the preference weight. Usually, a conditioned MORL model generates actions based on the given preference weight \cite{abels2019dynamic,yang2019generalized,kallstrom2019tunable}. They do not need to know the user's specific preference but a sample from the preference weight space during training. Yang et al. proposed Envelop Q learning\cite{yang2019generalized}, that adopts vectorized update. At each training step, it samples preference weights randomly. However, rather than randomly sample preference for training,  enhancements were conducted on this algorithm where the weight that can achieve the largest improvement is sampled. Basaklar et al. \cite{basaklar2022pd} proposed a method to efficiently update the model based on the angle between the Q-values and the weight vectors. Alegre et al. \cite{alegre2023sample} leveraged \textit{GPI} to find an update direction with the largest potential to propel the improvement of the whole policy set.

\subsection{MORL for Energy Management}
\label{subsec: residential energy management}
MORL has been widely used in energy management, e.g. micro-grid control \cite{xu2022preference,liu2023multiobjective}, water heating system oversight \cite{riebel2024multi}, energy control in vehicles \cite{zheng2022reinforcement,wu2023multi}, and the management of residential energy systems \cite{lu2022multi,lu2023go,lu2024inferring}. In assessing multi-objective energy management tasks, linear scalarization is usually employed to determine the aggregated return from various objectives \cite{hayes2022practical}. The weights reflecting the potential preferences of users are on a simplex. As the number of objectives increases, the complexity of the problem escalates exponentially. A noteworthy consideration is that a finite number of weight factor combinations are available \cite{roijers2016multi}. This has prompted the application of fuzzy logic techniques to streamline the solution set in MORL for energy management \cite{conti2012optimal, pourghasem2019stochastic,saberi2019optimal,yang2021multi}. Liu et al. \cite{liu2023multiobjective} introduced a policy-based, model-free MORL algorithm that employs the Borg MOEA approach \cite{hadka2013borg} for policy improvement. Despite these advancements, a value-based evolutionary MORL algorithm remains a gap. Although an actor-critic model of evolutionary reinforcement learning is proposed \cite{khadka2018evolutionary}, it is tailored for single-objective scenarios and has not been adapted for MORL contexts. Moreover, the model-based MORL algorithm \cite{alegre2023sample} is not evaluated within residential energy management yet.

Wu et al. have used a prioritized dueling double DQN to formulate a multi-objective energy management system, integrating multiple objectives within a singular value function to optimize cumulative rewards. This strategy, similar to the approaches adopted by Riebel et al. \cite{riebel2024multi} and Xu et al. \cite{xu2022preference}, involves incorporating various objectives into one reward function. This potentially results in significant computational demands when the preference for the objectives changes. Cutting-edge MORL algorithms can dynamically adjust to changes in the weight combination of objectives \cite{alegre2023sample}, yet their practical evaluation in the context of residential energy management remains unexplored.

\subsection{Meta Learning}
\label{subsec: continual reinforcement learning}
Humans can get satisfactory performance on a new task with a few attempts if they have some related knowledge. Although AI players can reach the human levels in many cases, they always need more samples. Meta-learning was proposed to train a model to quickly adapt to new scenarios by using a small amount of data\cite{finn2017model}. Vettoruzzo et al. conducted a comprehensive review of the meta-learning technologies. They provide a taxonomy of meta-learning methods, i.e. black-box meta-learning methods; optimization-based meta-learning methods; meta-learning via distance metric learning; and hybrid approaches \cite{vettoruzzo2024advances}.

In this work, we concentrate on optimization-based meta-learning methods. Consequently, while hybrid approaches, such as those discussed in \cite{wang2019hybrid}, offer valuable insights and advancements by integrating various meta-learning strategies, they extend beyond the scope of our current analysis. 

A foundational work in optimization-based meta-learning is \textit{Model-Agnostic Meta-Learning (MAML)} \cite{finn2017model}, renowned for its broad applicability and impact. \textit{MAML} learns initial model parameters which can fast adapt to a new task with only few-shot training. There is an inner loop and an outer loop in \textit{MAML}. In the inner loop (fast adaptation), for each sampled task, the current model parameters are used to conduct one or several gradient descent steps. This is to obtain the task-optimized parameters. In the outer loop (meta-update), the performance across all tasks is evaluated. This is used to back-propagate through the model's initial parameters for update. With these two loops, a set of initial parameters that allows the model to quickly adapt to new tasks is found.

Nevertheless, due to the computational burden associated with the bi-level gradient optimization in \textit{MAML}, algorithms like \textit{Reptile} and \textit{FOMAML} \cite{nichol2018first} have been developed to alleviate these issues. \textit{Reptile} seeks to identify a set of initial parameters that are near the optimal parameters for individual tasks. In our research, we employ \textit{Reptile} as the meta-learning methodology due to its computational efficiency and its compatibility with neural network architectures.

\section{Preliminary}
\label{sec: preliminary}
We model the multi-objective residential energy management problem as a \textit{multi-objective Markov decision process} (MOMDP), i.e. $\mathcal{M}:=(\mathcal{S},\mathcal{A},\mathcal{T},\gamma,\mu,\bm{R})$ \cite{hayes2022practical}. 
The spaces of states and actions are denoted as $\mathcal{S}$ and $\mathcal{A}$; $\mathcal{T}:S\times A\times S \xrightarrow[]{} [0,1]$ is the probabilistic transition function; $\gamma\in[0,1)$ is a discount factor; $\mu:S_{0}\xrightarrow[]{}[0,1]$ is the initial state probability distribution; The function $\bm{R}: \mathcal{S} \times \mathcal{A} \times \mathcal{S} \rightarrow \mathbb{R}^d$ is a vectorized reward function, focusing on two main objectives in this work: maximizing comfort and minimizing energy costs. We start with introducing some necessary concepts of MORL setting.

In MORL setting, the \textit{value vector}, starting from an initial state distribution $\mu$ then following policy $\pi$ is $\bm{v}^{\pi}:=\mathbb{E}_{s_{0}\sim\mu}[\bm{q}^{\pi}(s_{0},\pi(s_{0})]$.The $i$-th component of $\bm{v}^{\pi}$ represents the value returned for the $i$-th objective. With the value vector of policy, we can define the Pareto dominance relation ($\succ_{p}$): $\bm{v}^{\pi}\succ_{p}\bm{v}^{\pi'}\iff(\forall i:v_{i}^{\pi}\geq v_{i}^{\pi'})\cap (\exists i:v_{i}^{\pi}>v_{i}^{\pi'})$ 
We say that $\bm{v}^{\pi}$ is non-dominated when at least one element of $\bm{v}^{\pi}$ is greater than all other $\bm{v}^{\pi'}$. The Pareto front (PF) is therefore defined: $\mathcal{F}:=\{\bm{v}^{\pi}|\nexists \pi'\ s.t.\ \bm{v}^{\pi'}\succ_{p}\bm{v}^{\pi}\}$

In MORL, policy evaluation is dependent on the preference vector given. To involve different user-defined preferences over the objectives, a \textit{utility function} is used \cite{hayes2022practical} to scalarize the reward vector. Linear utility function \cite{hayes2022practical} is the most frequently used utility function. A linear weight vector $\bm{w}$ denoting the user preference of importance over each objective is given to scalarize $\bm{v}^{\pi}$.
\begin{equation}
    u(\bm{v}^{\pi},\bm{w})=v_{\bm{w}}^{\pi}=\bm{v}^{\pi}\cdot{\bm{w}}
\end{equation}
where $\bm{w}$ from the simplex $\mathcal{W}: \sum_{i}^{d}w_{i}=1, w_{i}\geq0$. The convex coverage set, \textit{CCS} represents a finite convex selection of the \textit{Pareto front (PF)}\footnote{Under linear utility function the $CCS$ is equivalent to the \textit{PF}}. Each point in the \textit{CCS} is at least one optimal policy corresponding to a certain linear preference. Formally defined as: 

$CCS:=\{\bm{v}^{\pi}\in\mathcal{F}|\exists\bm{w}\ s.t.\ \forall \bm{v}^{\pi'}\in\mathcal{F},\bm{v}^{\pi}\cdot\bm{w}\geq\bm{v}^{\pi'}\cdot\bm{w}\}$.

When using MORL in residential energy management, as renewable energy source is introduced, different weather conditions over the year can cause very different generation backgrounds. Traditional appoaches often struggle to cope with such environmental variability. \footnote{In this work the weather data is from a London household \cite{lawrie2022} and the resultant renewable generation is from the simulation.} Meta-learning aims at fast adaptation to new tasks through training on a variety of tasks, which aligns seamlessly with our scenario. The next challenge is to recognize when the underlying environment changes. We use the unsupervised anomaly detection technique to distinguish different renewable generation contexts as distinct tasks and utilize a meta-learning approach to identify initial model parameters that are suitable for rapid fine-tuning during evaluation. 

Our concern is to enhance the \textit{GPI}-based approach with the meta-learning method to initialize the model with a set of parameters that performs generally well in all contexts and can be effectively finetuned through a few-shot manner.

\section{R-GPI-LS/PD Algorithm} 
\label{sec: algorithm}
We first introduce the method we used to do context shift detection. Then we outline the meta-learning \textit{GPI} algorithms. 

\subsection{Context Detection}
\label{subsec:context detection}
In this work, the context shifting points in residential energy management are defined as the points in time when changes in weather cause significant variations in the power output of renewable energy sources (substantial changes in the MDP).
Those shifting points are not easily detectable, and there is also a lack of labeled data to support the supervised learning model. We leverage the unsupervised learning paradigm based on AE architecture \cite{audibert2020usad, yang2023adt, yang2024adt}. 

The AE is designed to try to reconstruct the input data. Such a model, when trained on data from a specific context can gain good reconstruction ability. We reversely use its design mechanism to detect the context-shifting points. When data from a different context comes, the reconstruction loss will see a significant rise. 

We first train the AE on an initial context data. Once the training converges (or even overfits), we proceed with the reconstruction of the subsequent data, while monitoring the AE's reconstruction loss. We average the loss values over a moving window. When the reconstruction loss for a specific context's data points exceeds the rolling average, it indicates that a significant shift in context has occurred\footnote{We define a variation as ``significant" when the reconstruction loss of the auto-encoder for the current day is greater than the average reconstruction loss calculated over the previous 7 days. For instance, transitioning from the continuous rainy weather of winter and spring to the clear skies of summer inevitably leads to changes in solar irradiance, resulting in a significant increase in the power output of renewable energy sources. This will result in a higher reconstruction loss and therefore create a context-shifting point. Such variations can also occur within the same season, especially in climates with variable weather.}. At this point, we designate the current context as a new context and continue training the AE on this segment of context. This process is repeated until the detection is complete for the entire year's data. See Algorithm \ref{alg: Context Detection} for more detail.

\begin{algorithm}
    \caption{Context Detection}
    \label{alg: Context Detection}
    \begin{algorithmic}[1]
    \STATE \textbf{Input} Windowed annual dataset $\{win_{1},...,win_{n}\}$; Initial AE model; Threshold value $\delta=-\infty$; Reconstruction loss  function $\mathcal{L}$; List of reconstruction loss $[\ \ ]_\mathcal{L}=[\ \ ]$ ; Last window of this context $win_{c}=win_{1}$; Context list $[\ \ ]_{context}=[\ \ ]$ 
    \FOR{$win_{i}$ in $\{win_{1},...,win_{n}\}$}
        \IF{$\mathcal{L}$($win_{i}$, AE($win_{i}$))>$\delta$}
            \STATE {Add $i$ to $[\ \ ]_{context}$}
            \STATE {Empty the reconstruction loss list $[\ \ ]_\mathcal{L}=[\ \ ]$}
            \STATE {Train AE with the data $\{win_{c},...,win_{i}\}$}
        \ENDIF
        \STATE {Add $\mathcal{L}$($win_{i}$, AE($win_{i}$)) to $[\ \ ]_\mathcal{L}$}
        \STATE {Update the threshold $\delta \xleftarrow{} mean({[\ \ ]_\mathcal{L}}$)}
    \ENDFOR
    \STATE \textbf{Output} $[\ \ ]_{context}$
    \end{algorithmic}
\end{algorithm}
\vspace{-20pt}
\subsection{Reptile and GPI Algorithm}
\label{subsec:Meta Learning Enhanced GPI}
To make the model able to adjust to non-stationary environments while saving computational overhead, we use the meta-learning paradigm to improve the \textit{GPI}-based algorithm. We use \textit{Reptile} \cite{nichol2018first} to do the meta-learning task. \textit{Reptile} is a first-order, gradient-based meta-learning algorithm that we favor due to its ability to avoid second-order gradient computations \cite{finn2017model} and save computational resources.

\subsubsection{R-GPI-LS/PD Meta Training}
\label{subsec:R-GPI-LS Meta Training}
The detail of the \textit{Reptile}-based \textit{GPI-LS/PD} (\textit{R-GPI-LS/PD}) meta-training process is shown in Algorithm \ref{alg: R-GPI}. After detecting the contexts with Algorithm \ref{alg: Context Detection}, at each epoch of meta-training, one context is sampled from the context list. A new set of parameters $\phi'$ is calculated after \textit{GPI} update. Then the parameter of the model is updated as done in line 6 in Algorithm \ref{alg: R-GPI}.

\begin{algorithm}
    \caption{R-GPI-LS/PD Meta Training}
    \label{alg: R-GPI}
    \begin{algorithmic}[1]
    \STATE {\textbf{Input} Learning rate $\epsilon$; Initial model $\phi$; Epochs number $n_{epochs}$;}
    \STATE {\textbf{Run} Algorithm \ref{alg: Context Detection} to find the set of different $[\ \ ]_{context}$}
    \FOR{$iteration=1$ to $n_{epochs}$}  
    
        \STATE Sample a $context$ from $[\ \ ]_{context}$
        \STATE Get updated parameter: $\phi'\xleftarrow{}$GPI$(\pi_{\phi,context})$
        \STATE Do update the original parameter: $\phi\xleftarrow{}\phi+\epsilon(\phi'-\phi)$
    \ENDFOR
    \STATE {\textbf{Output} Few-shot finetune MORL policy model}
    \end{algorithmic}
\end{algorithm}
Following the meta-training process, the model is updated to parameters that can achieve generally good performance and are easy to finetune on all tasks.

\subsubsection{Finetune R-GPI-LS/PD}
\label{subsec:R-GPI-LS Fine-tuning}
After the meta-training phase, the model is capable of being few-shot finetuned. The finetuning procedure for the \textit{R-GPI-LS/PD} model is elaborated in Algorithm \ref{alg: R-GPI Fine-tuning}. Initially, the meta-trained model, denoted as $\phi$, is finetuned using data from the first day. Subsequently, the policy continues to operate until a contextual shift is detected. Upon detection, the original $\phi$ is finetuned with data from the current day. This repeats throughout until the entire year's data is used up. During the finetuning process, the rewards obtained are recorded and subsequently summed at the end of the process to calculate the expected utility for the entire year.
\begin{algorithm}
    \caption{R-GPI-LS/PD Finetuning}
    \label{alg: R-GPI Fine-tuning}
    \begin{algorithmic}[1]
    \STATE {\textbf{Input} Learning rate $\alpha$; Pre-trained (with Algorithm \ref{alg: R-GPI}) model $\phi$; Windowed annual dataset $\{win_{1},...,win_{n}\}$;}
    \FOR{$win_{i}$ in $\{win_{1},...,win_{n}\}$}
        \IF {$win_{i}$ is recognized as a new context by Algorithm \ref{alg: Context Detection}}
        \STATE {Sample $win_{i}^{sub}\subsetneq win_{i}$}
        \STATE {$\phi_{i}\xleftarrow{}$GPI$(\pi_{\phi, win_{i}^{sub}})$ } \COMMENT{Few-shot finetuning}
        \ENDIF
        \STATE {Conduct $\pi_{\phi_{i}}$ in the environment}
    \ENDFOR
    \end{algorithmic}
\end{algorithm}
\vspace{-20pt}
\section{Experiments}
\label{sec: experiment}
In this section, we describe the simulation environment, which is constructed using real-world data. We detail the baselines, specifying the data volume and the number of training steps (interchangeably referred to as the training budget) allocated to each. We then introduce the metrics used, i.e. expected utility, PF approximation visualization, hypervolume, sparsity and the return vector for two specific cases either cares more about the cost or maximize the comfort.

\subsection{Experiment Settings}
\label{subsec: Experiment Settings}
\subsubsection{Benchmark Environments}
\label{subsubsection: benchmark environments}
The weather data and background power demand\footnote{The power demand of the other appliances manually turned on/off by the user.} data used in this study are derived from residential settings in London (Latitude 51.331, Longitude 0.033, Elevation 182.3m, Hourly-based); Specifically, the weather data is obtained from \cite{lawrie2022} and processed using SAM \cite{sam2022} to simulate renewable energy generation (Using Solar Panel of SunPower Performance 17 SPR-P17-335-COM). Residential energy consumption data comes from \cite{kelly2015uk}, and electricity pricing is sourced from \cite{moneysavingexpert2023economy7} with British Gas electricity rates at 36.62p/kWh during 08:00 - 23:00 and 15.18p/kWh from 23:00 - 08:00.

We have removed the solar heat pump and other related boilers from the house. Instead, we have integrated the \textit{Ariston VELIS EVO 80 L Electric Storage Water Heater} with a capacity of 1.5 KW, scheduling its operation for 4 hours between 0:00 - 8:00 every day. The agent manages the heater operation on an hourly basis.

The two objectives for MORL are: 

1. to save the cost (\pounds); 

2. to maximize the comfort (try the best to make the appliance to work in the assigned time slot.). 

The state space of our model is defined by a tuple consisting of the following elements: Background Power Demand (kW), Time (hrs), Remaining Task (hrs)\footnote{"Remaining Task" represents the number of hours left for an appliance to operate. For example, if an appliance needs to run for 4 hours every day and has been running for 1 hour, the remaining task is 3.}, and Renewable Generation (kW). 

The action space is binary, with only two options: 0 or 1. An action of "0" indicates turning the appliance off, whereas an action of "1" means turning the appliance on.

The reward space is constructed from two primary components: the hourly bill and comfort. The comfort reward is set to 1 when the appliance operates between 0:00 and 8:00 and the remaining task > 0, otherwise 0. The hourly bill reward is calculated as the negative value of the energy cost (\pounds), penalizing higher bill and encouraging energy-efficient behaviors. 

\subsubsection{R-GPI-LS/PD \& Finetune R-GPI-LS/PD}
We summarize our methods, i.e. \textit{R-GPI-LS/PD} and \textit{Finetune R-GPI-LS/PD}.

\textbf{i.} \textit{R-GPI-LS/PD}: It centers on meta-training the \textit{GPI-LS/PD} policy with daily data at contextual shift points, emphasizing the ability to swiftly adapt to significant context shifts. Through this process, the method establishes a set of initial parameters that are easy to finetune. 

\textbf{ii.} \textit{Finetune R-GPI-LS/PD}: With the initial parameters from R-\textit{GPI-LS/PD}. During the interaction with the environment, when new contextual shifts are detected, the policy is finetuned with the current day's data. This continuous finetuning is can further improve the model's performance throughout changing contexts over the year.

\subsubsection{Baselines}
\label{subsubsec:Baselines}
We detail the baselines we used in this Section. The primary baselines are variants of GPI-LS/PD as it is the current state of the art. Notably, other cutting-edge MORL algorithms such as SFOLS \cite{alegre2022optimistic} and Envelop Q Learning \cite{yang2019generalized} were compared with GPI-LS/PD in the work of \cite{alegre2023sample} and were found to be less effective. 

\textit{5.1.3.1 Baselines in Main Evaluation}

To evaluate of our methods, several baseline models are used:

\textbf{i.} \textit{GPI-LS/PD(month)}: It involves training a plain \textit{GPI-LS/PD} model on data collected from January 2014 for 40,000 steps.

\textbf{ii.} \textit{Finetuning GPI-LS/PD}:
Starting with \textit{GPI-LS/PD(month)}, it is retrained with the current day's data of  detected contextual shifts (12) for an additional 5,000 steps each.

\textbf{iii.} Rule-based policies: 
Rule 1: operates the appliance between 0:00 and 4:00.
Rule 2: operates the appliance between 4:00 and 8:00.

\textit{5.1.3.2 Baselines in Ablation Study}

To understand the contributions of various components of our method, we conduct an ablation study from two distinct perspectives:

\textbf{i.} Without the meta-learning method and context detection: 

\textit{GPI-LS/PD (year)}: A plain \textit{GPI-LS/PD} model is trained with the entire year data (2014-2015). This evaluates the performance of the \textit{GPI-LS/PD} model simply with much more train data and without any meta-learning or context-detection approach.

\textbf{ii.}  Without meta-learning method but with context detection:

\textit{Joint Training GPI-LS/PD}: It incorporates solely context detection to the plain \textit{GPI-LS/PD} to assess the impact of context detection on the model's adaptation to context shifts. We use a strong baseline i.e. joint training, which was also used in \cite{nichol2018first}.
The plain {GPI-LS/PD} model is trained with concatenated day-based data from the 12 shifting points. 
\textbf{iii.} The same rule-based policies as Section 5.1.3.1.


\subsubsection{Candidate Methods Setting}
We detail the training data volume and the training budget (the number of train steps) in Table \ref{tab:Number of Samples} to compare the data and train efficiency.

The data volume is determined by multiplying the number of days by the 24 hours. Specifically, the \textit{GPI-LS/PD (month)} and \textit{GPI-LS/PD (year)} models utilize 720 and 8,650 data samples, respectively. \textit{Joint Training GPI-LS/PD}, \textit{R-GPI-LS/PD} and \textit{Finetune R-GPI-LS/PD}, only use 288 data samples. \textit{Finetune GPI-LS/PD} uses 720+288=1008 data samples\footnote{\textit{Finetune R-GPI-LS/PD} uses the same data as \textit{R-GPI-LS/PD}.}.

\textit{GPI-LS/PD(month)}, \textit{Finetune GPI-LS/PD}, \textit{GPI-LS/PD(year)}, and \textit{Joint Training GPI-LS/PD} use substantially more training budget. \textit{Finetune GPI-LS/PD} are trained for 5000 steps at each shifting point. \textit{R-GPI-LS/PD} are trained through 3 iterations on 10 shifting points sampled out of the 12 at each iteration (each point  for 480 steps). \textit{Finetune R-GPI-LS/PD} are trained for 96 steps on each shifting point. However, the 14400 timesteps should also be counted as it is trained based on \textit{R-GPI-LS/PD}\footnote{The same for \textit{Finetune GPI-LS/PD}, it needs to add the steps used to train GPI-LS/PD(month).}. 

\begin{table}[htbp]
\caption{Data Volume and Training Budget}
\label{tab:Number of Samples}
\begin{center}
\fontsize{7}{2}\selectfont
\begin{tabular}{c|cc}
\toprule
Methods&Data Volume&Training Budget\\
\midrule
GPI-LS/PD(month)&30*24=720&40000\\
\midrule
Finetune GPI-LS/PD&15*24+720=1008&12*5000+40000=100000\\
\midrule
GPI-LS/PD (year)&365*24=8760&40000\\
\midrule
Joint Training GPI-LS/PD&12*24=288&40000\\
\midrule
R-GPI-LS/PD&12*24=288&10*3*480=14400\\
\midrule
Finetune R-GPI-LS/PD&12*24=288&14400+12*96=15552\\
\bottomrule
\end{tabular}
\end{center}
\end{table}
We have adjusted the number of gradient updates from 20 originally used \cite{alegre2023sample} to just 1, because the original frequency of gradient updates do not yield satisfactory results in our scenario. As each of the candidate algorithms uses the same gradient update frequency, this does not impact the fairness.


Each candidate method is executed multiple times using different random seeds to ensure the robustness of our results. The hyperparameters are illustrated in Appendix \ref{sec:Hyperparameter}.
\subsubsection{Evaluation Metric}
\label{subsubsec:metric}
All baselines are evaluated on the performance over a full year (2014-2015). We use multiple MORL metrics to evaluate our methods and baselines, i.e. expected utility (EU)\footnote{Calculated as $EU(\Pi)=E_{\textbf{w}\sim\mathcal{W}}[max_{\pi\in\Pi}v_{\textbf{w}}^{\pi}]$. This expectation is approximated by averaging the utility obtained from 100 evenly distributed weights from the simplex $\mathcal{W}$}, PF approximation visualization, hypervolume (HV)\footnote{The sum of the hypercube spanned by the reference point and the solutions on PF. We use [-1300, 0] as the reference point.}, and sparsity (Sp)\footnote{The sparsity $Sp(S)=\frac{1}{|S|-1}\sum^{m}_{j=1}\sum^{|S|}_{i=1}(\widetilde{S}_{j}(i)-\widetilde{S}_{j}(i+1))^{2}$, where $S$ is the PF approximation, $m$ is the number of objectives, $\widetilde{S}_{j}(i)$ is the $i$-th element of the sorted solutions on $j$-th objectives in $S$} of the PF solutions. For the specific evaluation for the energy community, we also evaluate the candidate methods by comparing the energy bill on preference [0.9,0.1] and comfort on preference [0.1,0.9].\footnote{The weight vector is the preference on [save cost, maximize comfort].} 

As emphasized in the work of Hayes et al. \cite{hayes2022practical}, EU is a preferable metric as it is more suitable to assess the solution's practical value for the user, while the HV is a bit problematic for comparing solutions in real-scenario especially when utility function is known to be linear, and Sp is the metric to measure the density of coverage of the whole PF as a alternates of HV. The solution set with a higher HV and lower Sp is preferable. Given the high level of user involvement in our scenario, we prioritize the evaluation metrics accordingly, placing the greatest emphasis on the EU, followed by HV and Sp together to reflect the user-centric assessment. Simply speaking, the higher the EU and HV and the ratio of HV/Sp are, the lower the Sp is, the better the solutions are. For the energy area specified two metrics, we say that the lower the bill is when given [0.9,0.1], and the higher the comfort when given [0.1,0.9], the better the solution set is. 

\section{Experiment Result and Discussion}
\label{subsec: Result}
This section consists of the result of context detection, the main evaluation, the ablation study, and the discussion and results summary. 

\subsection{Result of Context Shifting Detection}
\label{subsubsec:Result of Context Shifting Detection}
\begin{figure}[ht]
\begin{center}
\centerline{\includegraphics[width=\columnwidth]{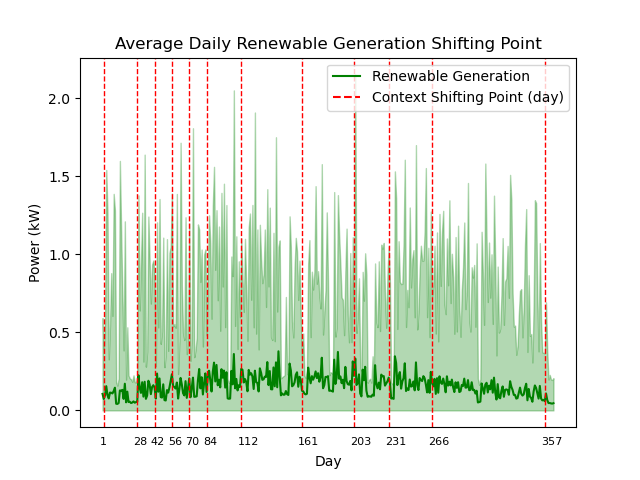}}
\caption{Context Shifting Detection}
\label{fig:Context_Shifting_Detection}
\end{center}
\end{figure}

We present the outcomes of detecting context-shifting points (12 shifting points on days 1, 28, 42, 56, 70, 84, 112, 161, 203, 231, 266, and 357.) in Fig. \ref{fig:Context_Shifting_Detection}.

One interesting observation is that the shifting points are more heavily concentrated towards the start of the year. This probably stems from the "spring cold snap", where spring temperatures suddenly drop to near-winter levels. This usually happens because cold air from the north can still affect warmer regions during spring and may cause sudden clouds, rain, or hail and influence renewable generation. Conversely, "autumn cold snap" is not common. This may be because the gradual temperature drop typically progresses steadily into winter without the sudden cooling seen in spring. It is therefore more stable than the spring season.


\subsection{Main Evaluation}
\label{subsubsec:Main Evaluation}
\begin{figure}[ht]
\begin{center}
\centerline{\includegraphics[width=\columnwidth]{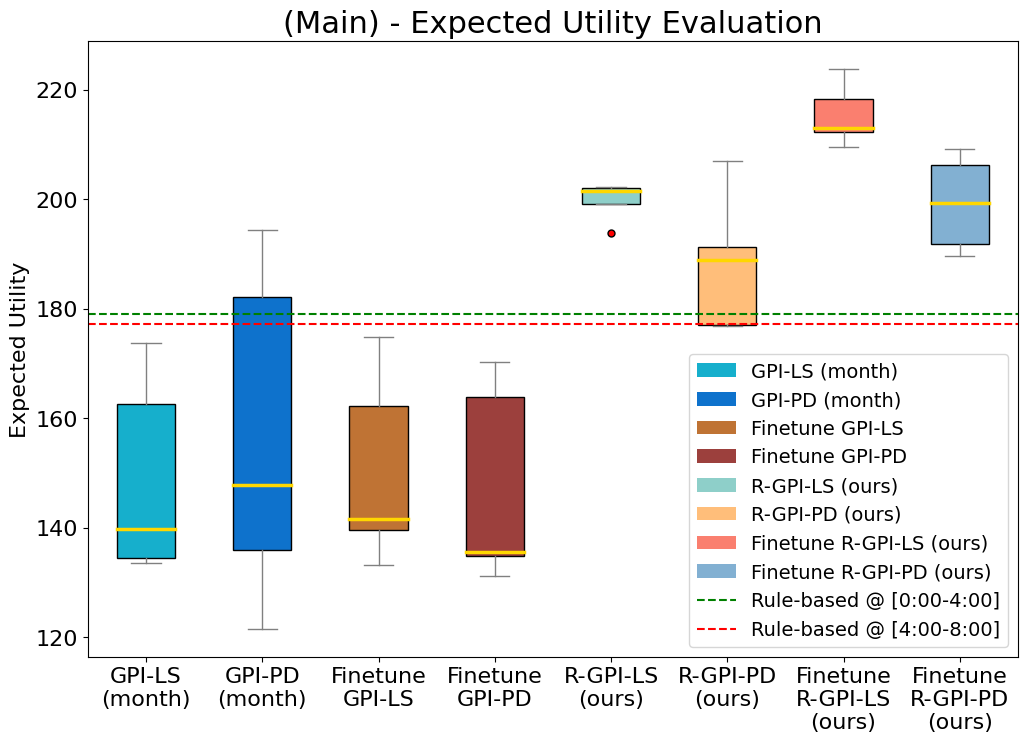}}
\caption{Main Evaluation Result - Expected Utility Evaluation - GPI\_LS}
\label{fig:Main_EU}
\end{center}
\end{figure}

Evaluation results on EU, HV, Sp, and HV/Sp are shown in Fig. \ref{fig:Main_EU}. The \textit{GPI-LS (month)} and \textit{Finetune GPI-LS} models display similar median values but the latter has a smaller variance. \textit{GPI-PD (month)}  has the highest median and max values among the first four baselines. It is counter-intuition that \textit{Finetune GPI-PD} is the worst among the four baselines as the environment model was supposed to help with the policy improvement. This may be because that the environment is hard to be accurately modelled during finetuning and therefore hinders the policy improvement. Unfortunately, neither of these four baselines' median exceeds the rule-based policy. This shows that the pure GPI-LS/PD struggle in such changing environments. 

Our candidates, \textit{R-GPI-LS/PD} and \textit{Finetune R-GPI-LS/PD} surpass all baselines except a small overlap between \textit{R-GPI-/PD} and \textit{GPI-PD (month)}. Although \textit{R-GPI-PD}'s performance sometimes falls below the rule-based policy, its median value is higher. After finetuning, \textit{R-GPI-LS/PD} surpasses other candidates. This demonstrates our methods' superior performance even with significantly less data (60.0\% - 71.43\% less) and a limited training budget (61.12\% - 85.6\% less). See Table \ref{tab:Comparison} in Appendix \ref{sec:Results Comparison} for more details of comparison. One interesting observation is that although \textit{R-GPI-LS/PD} receives performance improvement after finetuning, the model-based variants fail to surpass their model-free counterparts. This aligns with the observation between \textit{Finetune GPI-LS/PD} and implies that the environment model struggles to grab accurate dynamic information when the context is swiftly changing in finetune.

\begin{figure}[ht]
\begin{center}
\centerline{\includegraphics[width=\columnwidth]{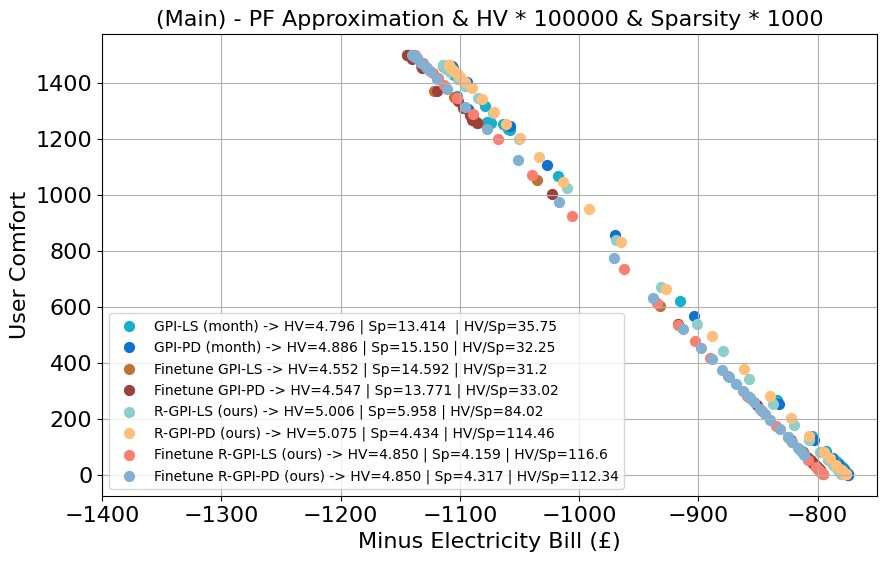}}
\caption{Main Evaluation Result- PF Approx. \& HV \& Sp}
\label{fig:Main_PF_HV_SP}
\end{center}
\end{figure}

In Fig. \ref{fig:Main_PF_HV_SP}, we visualize the PF approximation and evaluate the HV, Sp, and their ratio. The first four baselines evaluation shows that directly finetuning the learned policy does not have a significant influence on the performance. Though they can sometimes find better solutions than our methods, they still fail to achieve a better HV (except GPI-PD(month) that achieves a comparable HV as our methods) and lower sparsity. Our methods outperform all the baselines in the Sp, and HV/Sp. \textit{R-GPI-LS/PD} has achieved the highest HV and it has achieved the lowest Sp after fintuning. We observe that \textit{R-GPI-LS/PD} have seen a drop after finetune, however, their Sp also decrease, which demonstrates an improvement of the solution density/quality \cite{hayes2022practical}. During finetuning, the ratio of HV/Sp of our methods has either increased, or maintained at a comparable standard.

\vspace{-15pt}
\subsection{Ablation Study}
\label{subsubsec:Ablation Evaluation}
\begin{figure}[ht]
\begin{center}
\centerline{\includegraphics[width=\columnwidth]{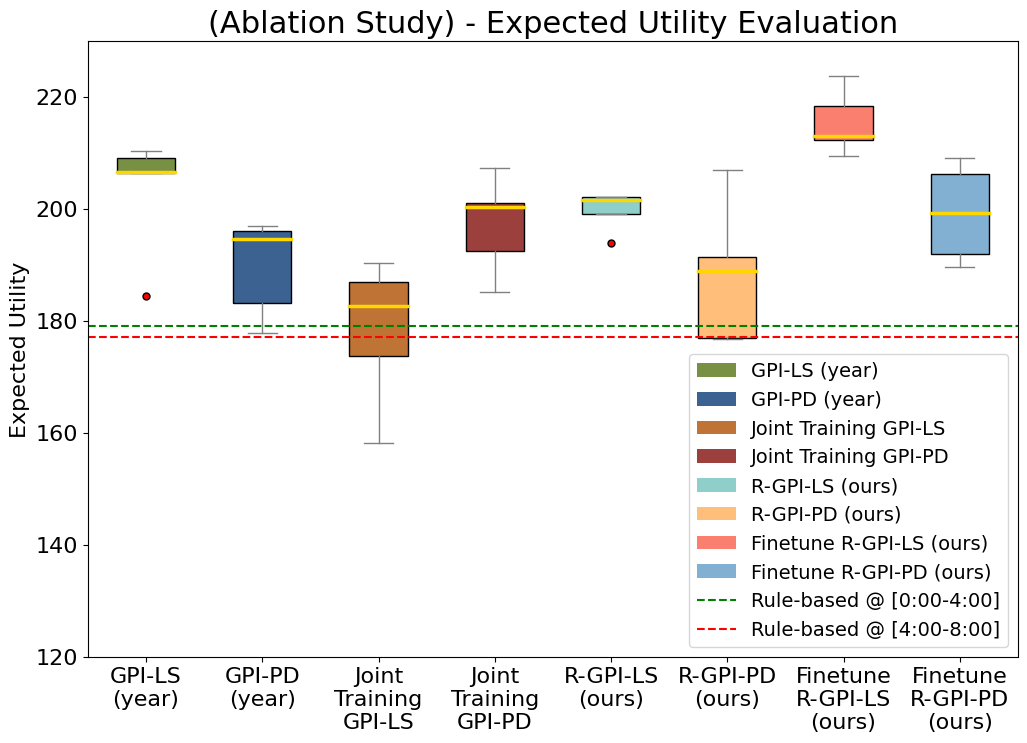}}
\caption{Ablation Study - Expected Utility Evaluation}
\label{fig:Ablation_EU}
\end{center}
\end{figure}

\begin{figure}[ht]
\begin{center}
\centerline{\includegraphics[width=\columnwidth]{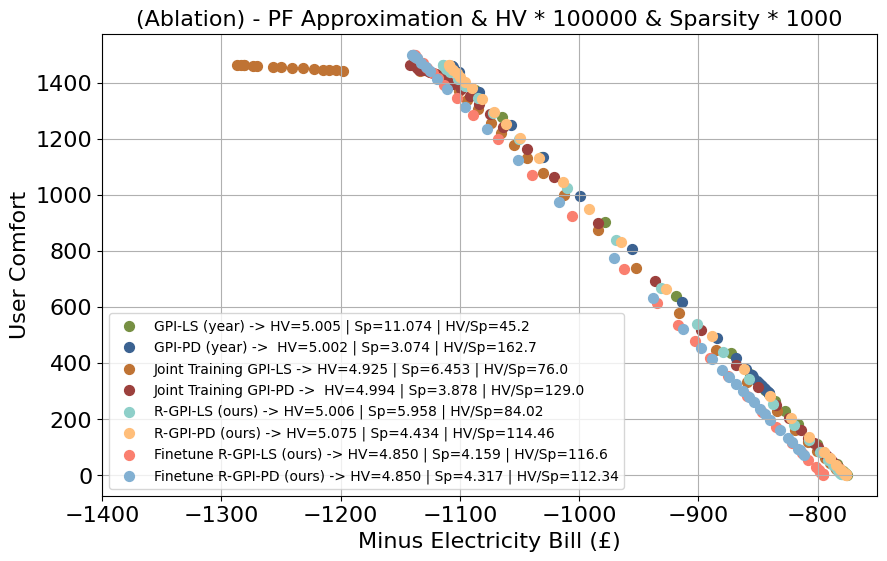}}
\caption{Ablation Study - PF Approx. \& Hypervolume \& Sparsity}
\label{fig:Ablation_PF_HV_Sp}
\end{center}
\end{figure}

We conduct the ablation study and show the results in Fig. \ref{fig:Ablation_EU} and Fig. \ref{fig:Ablation_PF_HV_Sp}. Recall that the ablation study focuses on two aspects, i.e. without both context detection and meta-learning, i.e. \textit{GPI-LS/PD (year)} and only without meta-learning, i.e. \textit{Joint Training GPI-LS/PD}. 

Each candidate's EU medians surpass rule-based policies. 
In most cases, \textit{GPI-PD} variants see larger variance than their \textit{GPI-LS} counterparts, except the joint training cases where the \textit{GPI-PD} variant is marginally lower in variance. This is consistent with Fig. \ref{fig:Main_EU} that \textit{GPI-PD} variants can bring more variance. This may stem from the fact that some state elements are outside the agent's control, e.g. renewable generation, which make the environment model's prediction of the next state incorrect and hinder the training. Notably, only the joint training setting and the month-based training setting have seen better performance from model-based variant. As our statement in last Section, i.e. "finetune process makes it hard to model the environment", we now further analyze this. The internal similarity of \textit{GPI-PD (month)} and \textit{Joint Training GPI-PD} are the training environment. \textit{GPI-PD (month)} is trained with a almost stable environment (1-30 days) where the next shifting point is just detected at day 28. \textit{Joint Training GPI-PD} is trained with the day-based concatenated 12 shifting points.  It is a synthetic environment and shifting points are evenly distributed. \textit{GPI-PD(year)} can be deemed as a data-dense version of joint training where the different contexts are also concatenated but not evenly distributed. However, the difference of context distribution between year-based train and joint training has seen contrary results. These observations mean that the model-based MORL algorithm needs either a stable environment or an environment that if the uncertainty is brought by states outside of the agent control it would be better that those context shifts are evenly distributed. We leave this for future work that the model-based RL algorithm needs improvement to handle these problems, i.e. unevenly distributed contexts, and performance drop in finetune.

In Fig. \ref{fig:Ablation_PF_HV_Sp}, we show the PF approximation, the HV, SP, and HV/SP. One interesting result is that the best performance on HV, SP, and HV/SP is \textit{GPI-PD(year)} which is not very well-performing according to the EU evaluation. This is because it finds some good solutions by following incorrect preferences. This can be proved by the result shown in Table \ref{tab:Summary}, where neither its bill or comfort is the best. It is consistent with the statement from \cite{hayes2022practical} that simply using HV or Sp cannot fully show the value of a solution set in a practical scenario. The second best method on HV/SP is \textit{Joint Training GPI-PD}, while our methods still are among the best-performing ones. 

The joint training methods have demonstrated better performance than \textit{GPI-LS/PD} or \textit{Finetune GPI-LS/PD} models, with 60\% and 71.43\% less data volume respectively. This indicates that context detection reduces the need for extensive data samples. The superior performance of our methods over the joint training further underscores the effectiveness of meta-learning, which achieves even better results with 64\% less training budget.

Although the year-based \textit{GPI-LS/PD} has comparable performance to other methods, it has a higher computational cost—using 96.71\% more data and over 60\% more training steps than our methods.

This comparison not only demonstrates the advantages of integrating meta-learning and context detection but also validates our highly efficient approach.

\subsection{Result Summary}
\label{subsubsec:Summary}
We summarize the evaluation in Table \ref{tab:Summary}\footnote{The up arrow $\uparrow$ means the higher the value is the better the policy is while the down arrow $\downarrow$ means the lower the value is the better the policy is. The bold value is the best performance on that metric}. See percentage comparison in Appendix \ref{sec:Results Comparison} and full solution visualization in Appendix \ref{sec:Full Solution Set}.

It does not surprise us that simply finetuning the plain \textit{GPI} models cannot achieve performance improvement (even cause decrease). This is because the internal representation of the trained model may only be suitable to specific contexts but not to others. When doing finetuning, these "stubborn" representations can hinder the parameter update. With meta-learning, the model can find the most context-sensitive parameters that are capable of later fintuning. 

The ablation study validates all components of our algorithms. Notably, \textit{GPI-LS/PD (year)} accesses diverse and more extensive data. Common sense suggests that training with more data should enhance model performance, however, this is not always true according to our experiment, i.e. the \textit{Joint Training GPI-PD} is better than \textit{GPI-PD (year)}. As we mentioned in Section \ref{subsubsec:Ablation Evaluation}, this should be due to the unevenly distributed context shifting.

\textit{Joint Training GPI-LS/PD} still are outperformed by \textit{Finetune R-GPI-LS/PD}, and merely match \textit{R-GPI-LS/PD}. As stated in \cite{nichol2018first}, that the joint-training method seeks to optimize on several tasks, while \textit{Reptile} considers second-and-higher order of derivatives when multiple gradient updates are conducted. This can explain the superior outcomes of our method.

Our \textit{Finetune R-GPI-LS} is the best performing algorithm when compared with the second best method \textit{GPI-LS(year)} where it uses 96.71\% less data and 61.1\% less training steps to achieve 5.9\% improvement on EU, 62.44\% improvement on Sp, 3.28\% on bill and 2.74\% improvement on comfort but only has 3.1\% drop on HV. We conduct a detailed comparison in Table \ref{tab:Comparison} in Appendix \ref{sec:Results Comparison}.
\begin{table}[htbp]
\caption{Summary of Evaluation}
\label{tab:Summary}
\begin{center}
\fontsize{6.3}{2}\selectfont
\begin{tabular}{c|ccccc}
\toprule
& \multicolumn{5}{c}{\textbf{Metric}}\\
\midrule
\textbf{Baselines}&\textbf{EU}$\uparrow$&\textbf{HV}$\uparrow$&\textbf{Sp}$\downarrow$&\textbf{Bill}$\downarrow$&\textbf{Comfort}$\uparrow$\\
\midrule
GPI-LS(month)&148.85&479597.66&13414.23&\pounds 781.96& 1296.2\\
\midrule
GPI-PD(month)&156.36&488625.97&15149.91&\pounds \textbf{775.03}& 1347\\
\midrule
Finetune GPI-LS&150.31&455233.65&14592.21&\pounds 802.41& 1325.6\\
\midrule
Finetune GPI-PD&147.10&454665.54&13770.9&\pounds 802.32& 1324.8\\
\midrule
GPI-LS(year)&203.37&500506.12&11073.96&\pounds 840.93& 1460\\
\midrule
GPI-PD(year)&189.72&500212.43&6453.05&\pounds 853.97& 1460\\
\midrule
Joint Training GPI-LS&178.37&492483.88&\textbf{3074.39}&\pounds 834.31& 1449\\
\midrule
Joint Training GPI-PD&197.29&499427.29&3878.36&\pounds 807.7& 1457.8\\
\midrule
R-GPI-LS&199.75&500631.35&5958.15&\pounds 791.39& 1464\\
\midrule
R-GPI-PD&188.2&\textbf{507535.27}&4434.33&\pounds 781.08& 1464\\
\midrule
Finetune R-GPI-LS&\textbf{215.37}&485013.61&4159.47&\pounds 813.35&\textbf{1500}\\
\midrule
Finetune R-GPI-PD &199.23&484968.16&4316.89&\pounds 857.89& \textbf{1500}\\
\bottomrule
\end{tabular}
\end{center}
\end{table}
\vspace{-30pt}

\section{Conclusion}
\label{sec: conclusion}
In this paper, we highlighted the suitability of meta-learning for energy management and demonstrate the superior performance by extending state-of-the-art MORL algorithms with \textit{Reptile}. We applied and evaluated the \textit{GPI-LS/PD} in residential appliance scheduling. 

There are several interesting questions identified in our research that merit further investigation:

1. We notice that finetuning technique cannot seamlessly fit model-based MORL. This was once mentioned in the work of Mendonca et al. \cite{mendonca2020meta} however, it is not yet fully explored in MORL settings. It remains a open-question that how to finetune the environment model to positively help with the training of MORL policy when the task contexts are not evenly distributed.

2. The environment used is only 2-objective, evaluations in the environment with more objectives (e.g. peak shaving) are still an open question. We also plan to do further investigation on our algorithm about the performance in other environments.

3. Our work only talks about single-agent cases, however, the real-life scenario usually involves multiple appliances for scheduling. This renders the problem as a multi-agent problem which needs further exploration.




\begin{ack}
This research is funded by Irish Research Council, the Government of Ireland
Postgraduate Scholarship (GOIPG/2022/2140).
\end{ack}



\bibliography{mybibfile}
\newpage
\appendix
\onecolumn
\section{Results Comparison}
\label{sec:Results Comparison}
Table \ref{tab:Comparison} shows the detailed comparison among our methods and the baselines on all metrics including data volume and training budget we used in this work. Similar to Table \ref{tab:Summary}, the symbol $\uparrow$ indicates that a higher value corresponds to better performance, while $\downarrow$ signifies that a lower value denotes superior performance. The comparison is conducted by 
\begin{equation}
    improvement=\frac{metric(candidate\_method)-metric(baseline)}{metric(baseline)}\cdot100\%
\end{equation}
\begin{table*}[htbp]
\caption{Percentage Improvement Comparison }
\label{tab:Comparison}
\begin{center}
\fontsize{7}{2}\selectfont
\begin{tabular}{c|ccccccc}
\toprule
& \multicolumn{7}{c}{\textbf{R-GPI-LS}}\\
\midrule
\textbf{Baselines}&\textbf{EU}$\uparrow$&\textbf{HV}$\uparrow$&\textbf{Sp}$\downarrow$&\textbf{Bill}$\downarrow$&\textbf{ Comfort}$\uparrow$&\textbf{Data Volume}$\downarrow$&\textbf{Training Budget}$\downarrow$\\
\midrule
GPI-LS/PD(month)&34.2\%\ /\ 27.76\%&4.39\%\ /\ 2.46\%  &-55.58\%\ /\ -60.67\%  &1.21\%\ /\ 2.11\%&12.95\%\ /\ 8.69\%&-60.0\%&-64.0\% \\

\midrule
Finetune GPI-LS/PD&32.89\%\ /\ 35.79\%&9.97\%\ /\ 10.11\%  &-59.17\%\ /\ -56.73\%  &-1.37\%\ /\ -1.36\%&10.44\%\ /\ 10.51\%&-71.43\%&-85.6\%\\

\midrule
GPI-LS/PD(year)&-1.77\%\ /\ 5.29\%&0.03\%\ /\ 0.08\%  &-46.2\%\ /\ -7.67\%  &-5.89\%\ /\ -7.33\%&0.27\%\ /\ 0.27\%&-96.71\%& -64.0\%\\

\midrule
Joint Training GPI-LS/PD&11.99\%\ /\ 1.25\%&1.65\%\ /\ 0.24\%  &93.8\%\ /\ 53.63\%  &-5.14\%\ /\ -2.02\%&1.04\%\ /\ 0.43\%&0.0\%&-64.0\%\\

\midrule
R-GPI-LS/PD&0.0\%\ /\ 6.14\%&0.0\%\ /\ -1.36\%  &0.0\%\ /\ 34.36\%  &0.0\%\ /\ 1.32\%&0.0\%\ /\ 0.0\%&0.0\%&0.0\%\\

\midrule
Finetune R-GPI-LS/PD&-7.24\% \ /\ 0.26\%&3.22\%\ /\ 3.23\%  &43.24\%\ /\ 38.02\%  &-2.7\%\ /\ -7.75\%&-2.4\%\ /\ -2.4\%&0.0\%&-7.41\%\\
\midrule
& \multicolumn{7}{c}{\textbf{R-GPI-PD}}\\
\midrule
GPI-LS/PD(month)&26.43\%\ /\ 20.36\%&5.83\%\ /\ 3.87\%&-66.94\%\ /\ -70.73\%&-0.11\%\ /\ 0.78\%&12.95\%\ /\ 8.69\%&-60.0\%& -64.0\%\\

\midrule
Finetune GPI-LS/PD&25.2\%\ /\ 27.93\%&11.49\%\ /\ 11.63\%&-69.61\%\ /\ -67.8\%&-2.66\%\ /\ -2.65\%&10.44\%\ /\ 10.51\%&-71.43\%&-85.6\%\\

\midrule
GPI-LS/PD(year)&-7.46\%\ /\ -0.81\%&1.4\%\ /\ 1.46\%&-59.96\%\ /\ -31.28\%&-7.12\%\ /\ -8.54\%&0.27\%\ /\ 0.27\%&-96.71\%&-64.0\%\\

\midrule
Joint Training GPI-LS/PD&5.51\%\ /\ -4.61\%&3.06\%\ /\ 1.62\%&44.23\%\ /\ 14.34\%&-6.38\%\ /\ -3.3\%&1.04\%\ /\ 0.43\%&0.0\%&-64.0\%\\

\midrule
R-GPI-LS/PD&-5.79\%\ /\ 0.0\%&1.38\%\ /\ 0.0\%&-25.58\%\ /\ 0.0\%&-1.3\%\ /\ 0.0\%&0.0\%\ /\ 0.0\%&0.0\%&0.0\%\\

\midrule
Finetune R-GPI-LS/PD &-12.62\%\ /\ -5.54\%&4.64\%\ /\ 4.65\%&6.61\%\ /\ 2.72\%&-3.97\%\ /\ -8.95\%&-2.4\%\ /\ -2.4\%&0.0\%&-7.41\%\\
\midrule
& \multicolumn{7}{c}{\textbf{Finetune R-GPI-LS}}\\
\midrule
GPI-LS/PD(month)&44.69\%\ /\ 37.74\%&1.13\%\ /\ -0.74\%&-68.99\%\ /\ -72.54\%&4.01\%\ /\ 4.94\%&15.72\%\ /\ 11.36\%&-60.0\%&-61.12\%\\

\midrule
Finetune GPI-LS/PD&43.28\%\ /\ 46.41\%&6.54\%\ /\ 6.67\%&-71.5\%\ /\ -69.8\%&1.36\%\ /\ 1.37\%&13.16\%\ /\ 13.22\%&-71.43\%&-84.45\%\\

\midrule
GPI-LS/PD(year)&5.9\%\ /\ 13.52\%&-3.1\%\ /\ -3.04\%&-62.44\%\ /\ -35.54\%&-3.28\%\ /\ -4.76\%&2.74\%\ /\ 2.74\%&-96.71\%&-61.12\%\\

\midrule
Joint Training GPI-LS/PD&20.74\%\ /\ 9.16\%&-1.52\%\ /\ -2.89\%&35.29\%\ /\ 7.25\%&-2.51\%\ /\ 0.7\%&3.52\%\ /\ 2.89\%&0.0\%&-61.12\%\\

\midrule
R-GPI-LS/PD&7.81\%\ /\ 14.44\%&-3.12\%\ /\ -4.44\%&-30.19\%\ /\ -6.2\%&2.77\%\ /\ 4.13\%&2.46\%\ /\ 2.46\%&0.0\%&8.0\%\\

\midrule
Finetune R-GPI-LS/PD &0.0\%\ /\ 8.1\%&0.0\%\ /\ 0.01\%&0.0\%\ /\ -3.65\%&0.0\%\ /\ -5.19\%&0.0\%\ /\ 0.0\%&0.0\%&0.0\%\\
\midrule
& \multicolumn{7}{c}{\textbf{Finetune R-GPI-PD}}\\
\midrule
GPI-LS/PD(month)&33.85\%\ /\ 27.42\%&1.12\%\ /\ -0.75\%&-67.82\%\ /\ -71.51\%&9.71\%\ /\ 10.69\%&15.72\%\ /\ 11.36\%&-60.0\%&-61.12\%\\

\midrule
Finetune GPI-LS/PD&32.55\%\ /\ 35.44\%&6.53\%\ /\ 6.66\%&-70.42\%\ /\ -68.65\%&6.91\%\ /\ 6.93\%&13.16\%\ /\ 13.22\%&-71.43\%&-84.45\%\\

\midrule
GPI-LS/PD(year)&-2.04\%\ /\ 5.01\%&-3.1\%\ /\ -3.05\%&-61.02\%\ /\ -33.1\%&2.02\%\ /\ 0.46\%&2.74\%\ /\ 2.74\%&-96.71\%&-61.12\%\\

\midrule
Joint Training GPI-LS/PD&11.69\%\ /\ 0.98\%&-1.53\%\ /\ -2.9\%&40.41\%\ /\ 11.31\%&2.83\%\ /\ 6.21\%&3.52\%\ /\ 2.89\%&0.0\%&-61.12\%\\

\midrule
R-GPI-LS/PD &-0.27\%\ /\ 5.87\%&-3.13\%\ /\ -4.45\%&-27.55\%\ /\ -2.65\%&8.4\%\ /\ 9.83\%&2.46\%\ /\ 2.46\%&0.0\%&8.0\%\\

\midrule
Finetune R-GPI-LS/PD &-7.49\%\ /\ 0.0\%&-0.01\%\ /\ 0.0\%&3.78\%\ /\ 0.0\%&5.48\%\ /\ 0.0\%&0.0\%\ /\ 0.0\%&0.0\%&0.0\%\\
\bottomrule
\end{tabular}
\end{center}
\end{table*}

Under most evaluations (not including mutual comparison among our methods), both \textit{R-GPI-LS/PD} and \textit{Finetune R-GPI-LS/PD} have achieved improvement on EU and HV. Just name a few comparisons between our methods and the well-performing baselines.

\textit{R-GPI-LS}, in comparison to \textit{GPI-LS (year)}, has demonstrated an improvement in Hypervolume (HV) by 0.03\% and a substantial reduction in Sparsity (Sp) by 46.2\%. Additionally, it achieved a 5.89\% reduction in electricity bills and a 0.27\% improvement in comfort. However, it experienced a decrease on EU by -1.77\%.

Similarly, \textit{R-GPI-PD}, when compared with \textit{GPI-LS (year)}, showed a notable improvement in HV by 1.4\% and a reduction in Sp by 59.96\%, along with a 7.12\% decrease in electricity bills and a 0.27\% improvement in comfort. However, it also recorded a decrease on EU by -7.46\%. \textit{R-GPI-LS/PD} were trained with 96.71\% fewer data and a 64\% less training budget than \textit{R-GPI-LS}, highlighting the efficiency of \textit{R-GPI-LS} in resource utilization despite some trade-offs in performance.

\textit{Finetune R-GPI-LS} is the best performing algorithm when compared with the second best method \textit{GPI-LS(year)} where it uses 96.71\% less data and 61.1\% less training steps to achieve 5.9\% improvement on EU, 62.44\% improvement on Sp, 3.28\% on bill and 2.74\% improvement on comfort but only has 3.1\% drop on HV.

The \textit{Finetune R-GPI-PD}, compared with the best-performing \textit{GPI-PD-based} baseline, i.e.  \textit{GPI-PD(year)}, while being trained with 61.12\% less steps has achieved 0.98\% improvement on EU and 2.89\% improvement on comfort, however also gotten a 2.9\% drop on HV, a 11.31\% drop on Sp and 2.83\% drop on bill.

This empirically proves that our method outperforms the baselines. Except \textit{Joint Training GPI-LS/PD}, our methods have achieved a significant reduction of the sparsity. Notably, our method after finetuning usually underperforms in saving bills than most of the baselines. This denotes that the policy is more "active" in interacting with the environment by a higher probability of turning on the appliance.

\section{Full Solution Set}
\label{sec:Full Solution Set}
\begin{figure}[ht]
\begin{center}
\centerline{\includegraphics[width=0.75\columnwidth]{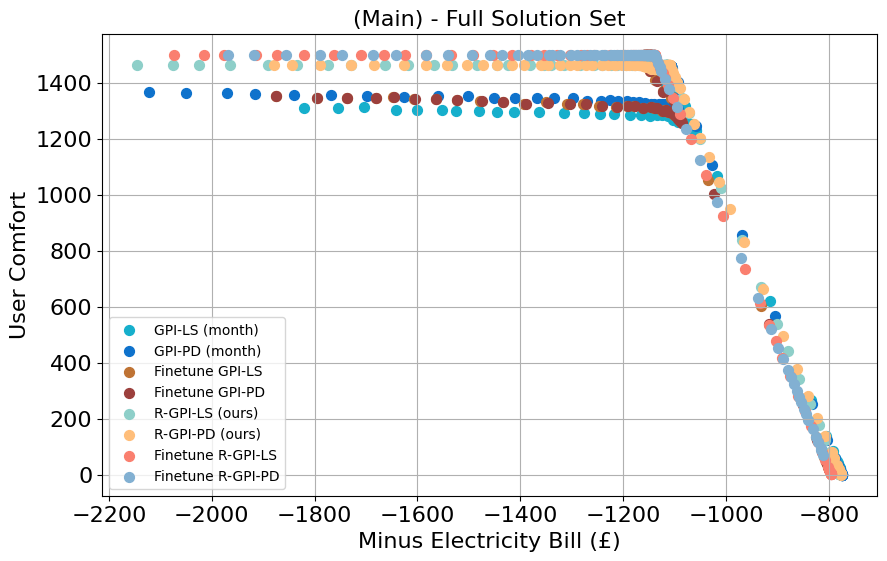}}
\caption{Main Evaluation - Full Solution Set}
\label{fig:Main_Full}
\end{center}
\end{figure}

\begin{figure}[ht]
\begin{center}
\centerline{\includegraphics[width=0.75\columnwidth]{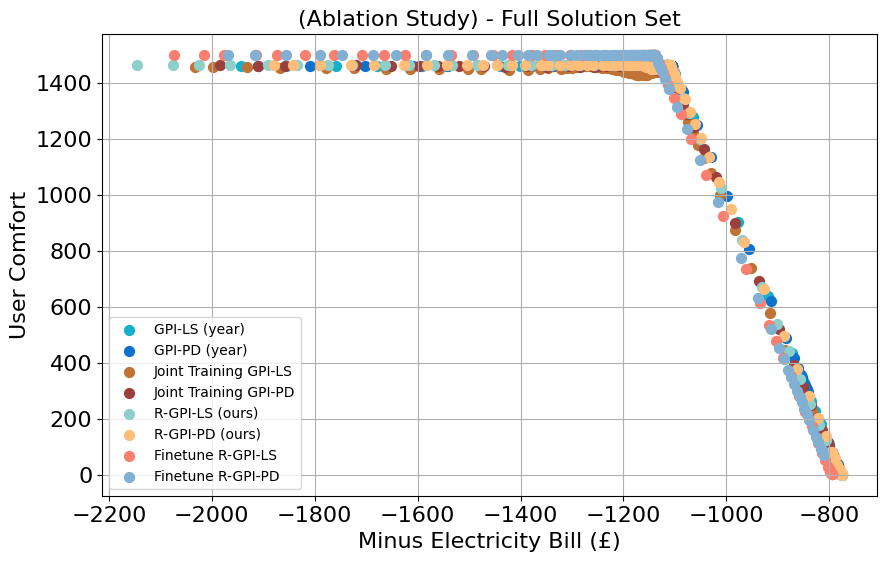}}
\caption{Ablation Study - Full Solution Set}
\label{fig:Ablation_Full}
\end{center}
\end{figure}

We show the full solution set rather than the PF of the methods. The full solution set of the main evaluation is shown in Fig. \ref{fig:Main_Full} while Fig. \ref{fig:Ablation_Full} shows the full solution set from the ablation study. 

It demonstrates that our methods are better at maximizing user comfort while they are slightly underperformed in bill cutting. Another observation is that for \textit{GPI-LS/PD(month)} and \textit{Finetune GPI-LS/PD}, there is a bump in the middle of the solution set, as shown in Fig. \ref{fig:Main_Full}, this means that these methods misaligned the solution with preference. In Fig. \ref{fig:Ablation_Full} and the result of our methods, the solution set is smoother and no obvious bump is seen, this means the policy can figure out the preference and make consistent decisions which is further reflected in the Sp evaluation, and also support the statement that HV is problematic from \cite{hayes2022practical}. The points in Fig. \ref{fig:Ablation_Full} are more condensed than the ones in Fig. \ref{fig:Main_Full}. This demonstrates that either with a larger data volume and training budget or by using context detection can the policy be effectively improved.

\section{Hyperparameters}
\label{sec:Hyperparameter}
We put the hyperparameters of the candidate method and the context detection model in Table \ref{tab:Hyperparameters}.
\begin{table*}[htbp]
\caption{Hyperparameters}
\label{tab:Hyperparameters}
\begin{center}
\fontsize{8}{10}\selectfont
\begin{tabular}{c|ccccc}
\toprule
Models& \multicolumn{5}{c}{\textbf{Hyperparameters}}\\
\midrule
&Learning Rate&Batch Size& Network Architecture& Target Update Frequency&Outer Loop Learning Rate \\
\midrule
GPI-LS/PD(month)&$3*10^{-4}$&$2*10^{5}$&[256, 256, 256, 256]&200&--\\

\midrule
Finetune GPI-LS/PD&$3*10^{-4}$&$2*10^{5}$&[256, 256, 256, 256]&200&--\\

\midrule
GPI-LS/PD(year)&$3*10^{-4}$&$2*10^{5}$&[256, 256, 256, 256]&200&--\\

\midrule
Joint Training GPI-LS/PD&$3*10^{-4}$&$2*10^{5}$&[256, 256, 256, 256]&200&--\\

\midrule
R-GPI-LS/PD&$3*10^{-4}$&$2*10^{5}$&[256, 256, 256, 256]&200&$3*10^{-4}$\\

\midrule
Finetune R-GPI-LS/PD&$3*10^{-4}$&$2*10^{5}$&[256, 256, 256, 256]&200&--\\
\midrule
&Learning Rate&Num of Epochs& Network Architecture& &\\
\midrule
Context Detection Model&$1*10^{-3}$&500&[24, 64, 32, 16]&--&--\\
\bottomrule
\end{tabular}
\end{center}
\end{table*}
\end{document}